\documentclass{IEEEtran}
\usepackage{cite}
\usepackage{amsmath,amssymb,amsfonts}
\usepackage{algorithmic}
\usepackage{graphicx}
\usepackage{textcomp}
\usepackage{multirow}
\usepackage{url}
\usepackage{bm}
\usepackage{multirow}
\usepackage{multicol}
\usepackage[ruled,longend]{algorithm2e}
\DeclareMathOperator*{\argmin}{argmin}

\def\BibTeX{{\rm B\kern-.05em{\sc i\kern-.025em b}\kern-.08em
    T\kern-.1667em\lower.7ex\hbox{E}\kern-.125emX}}
\begin{document}
\title{Exploring the Robustness of NMT Systems to Nonsensical Inputs}
\author{Akshay Chaturvedi$^{*}$, Abijith KP and Utpal Garain, \IEEEmembership{Member, IEEE}
\thanks{$^{*}$ Corresponding author. Email: akshay91.isi@gmail.com }
\thanks{Akshay Chaturvedi, Abijith KP and Utpal Garain are with Computer Vision and Pattern Recognition Unit, Indian Statistical Institute, India.}}

\maketitle

\begin{abstract}
Neural machine translation (NMT) systems have been shown to give undesirable translation when a small change is made in the source sentence. In this paper, we study the behaviour of NMT systems when multiple changes are made to the source sentence. In particular, we ask the following question ``Is it possible for an NMT system to predict same translation even when multiple words in the source sentence have been replaced?". To this end, we propose a soft-attention based technique to make the aforementioned word replacements. The experiments are conducted on two language pairs: English-German (en-de) and English-French (en-fr) and two state-of-the-art NMT systems: BLSTM-based encoder-decoder with attention and Transformer. The proposed soft-attention based technique achieves high success rate and outperforms existing methods like HotFlip by a significant margin for all the conducted experiments. The results demonstrate that state-of-the-art NMT systems are unable to capture the semantics of the source language. The proposed soft-attention based technique is an invariance-based adversarial attack on NMT systems. To better evaluate such attacks, we propose an alternate metric and argue its benefits in comparison with success rate.
\end{abstract}


\section{Introduction}

Neural machine translation (NMT) systems, with the advent of Transformers~\cite{transformer}, have achieved remarkable success in the past few years. Unlike recurrent architectures, Transformers are composed solely of \emph{attention layers} which allows for training at a lower cost (FLOPs). Moreover, Transformers have been shown to perform better than recurrent architectures. Recently, BERT-based embeddings~\cite{bert} have been used on a variety of natural language processing (NLP) tasks like question answering, natural language inference etc., achieving state-of-the-art results. In the field of computer vision, CNN-based systems despite achieving impressive performance have been shown to have \emph{blind spots} making them vulnerable to adversarial attacks. Given such a widespread usage of Transformer architecture, a natural question arises ``\emph{Are Transformers robust to noise in the input text or do they have blind spots as well?}".

In this regard, several studies have been done to study the robustness of NLP systems. Feng \emph{et al.} \cite{D18-1407} show that question answering models predict the same answer even when most of the words are removed from the question. To fool a text classifier, Ebrahimi \emph{et al.}\cite{hotflip} introduce HotFlip technique which shows that a character-level NLP model changes its prediction when few characters are flipped. In the ideal scenario, we want a character-level NLP model to be invariant to character flips (especially when the number of flips are few). Hence, it is undesirable if the prediction of a model changes when few characters in the input text are flipped. Belinkov and Bisk \cite{belinkov2018synthetic} show that NMT systems can be fooled via synthetic and natural character level noises. The aforementioned HotFlip technique has also been extended to neural machine translation (NMT) systems~\cite{hotflip-mt}. However in~\cite{hotflip-mt}, the authors study the robustness of a CNN-based recurrent architecture \cite{costa-jussa-character} rather than Transformer. With regards to robustness of Transformer, recent studies \cite{cheng-etal-2018-towards,cheng-etal-2019-robust} have shown that Transformers predict different translations for two semantically similar source sentences.

\begin{table}
\begin{center}
\small
\begin{tabular}{ll}
  \hline
  \textbf{src} & Not a single body should remain  \\
               &  undiscovered or unidentified . \\
  \hline
  \textbf{adv-src} & unaware topic single body should remain  \\
                   &  undsubmitted covered Within uniunclear \\
                   &  fied surely \\ 
  \hline
  \textbf{pred} & Kein einziger K\"orper sollte unbehandelt \\
                &  oder gekl\"art bleiben . \\
  \hline
\end{tabular}
\end{center}
\caption{Example of the proposed method. The English-German Transformer predicts the same translation for the two sentences even though multiple replacements are made.}
\label{table:ex}
\end{table}


In this paper, we explore the following question \emph{``Is it possible for an NMT system to predict same translation even when multiple words in the source sentence have been replaced?"}. This is different from the work of Cheng \emph{et al.}\cite{cheng-etal-2018-towards,cheng-etal-2019-robust}, where one expects the NMT system to predict the same translation. We perform the experiments on subword-level based NMT systems, namely BLSTM-based encoder-decoder with attention~\cite{D15-1166} and Transformer~\cite{transformer}. Given a source sentence $s=(s_{1}, s_{2},..., s_{n})$, our goal is to replace multiple words $s_{i}$'s with new words $s'_{i}$'s while ensuring that the predicted translation remains unchanged. To achieve this, we propose a soft-attention based technique.  Table~\ref{table:ex} shows one such example of the proposed technique. Since the NMT model is subword-level, in our experiments, we replace subword with word. For example, in Table~\ref{table:ex}, the word ``undiscovered" is broken into three subwords: und, is and covered. The subword ``is" is replaced by the word ``submitted" leading to the phrase ``undsubmitted covered" in the adversarial source sentence. In contrast to character flips, we want the NMT system to be sensitive (i.e. not invariant) to word flips, especially when multiple words are replaced. Such word-level \emph{invariances} captured by the model are \emph{undesirable}. The proposed technique might generate sentences which are semantically incorrect as shown in Table~\ref{table:ex}. Even in such cases, the NMT system is expected to give different translations since it may lead to lack of trust of the end user on the NMT system if two completely different sentences are assigned the same translation. This is in line with the work done by He and Glass \cite{he2018detecting} where a dialogue generator is expected to never output egregious sentences regardless of the \emph{semantic correctness} of the input sentence.

\subsection{Related work}

Several attempts have been made to study the robustness of NMT systems to noise in the input text~\cite{belinkov2018synthetic,hotflip-mt,cheng-etal-2018-towards,cheng-etal-2019-robust,liu-etal-2019-robust}. Ebrahimi \emph{et al.} \cite{hotflip-mt} propose HotFlip to attack a character-level NMT system. The HotFlip technique encodes a character flip as a vector. It chooses the optimum character flip based on the directional derivatives of the gradient of loss with respect to one-hot input along the \emph{flip vectors}. These directional derivatives give a first-order estimate of the change in loss when a character is flipped. This technique can be used for flipping words instead of characters as well. Cheng \emph{et al.} \cite{cheng-etal-2018-towards} show that replacing words by their synonyms lead to erroneous translations and propose an adversarial learning framework to ensure that the two source sentences, original and its noisy counterpart, get similar representations by the encoder of the NMT system. Cheng \emph{et al.} \cite{cheng-etal-2019-robust} show that the NMT systems output different translations for two semantically similar source sentences. The authors propose a training framework which uses the original training data along with the noisy data to enhance the robustness of NMT systems to such noise. Liu \emph{et al.} \cite{liu-etal-2019-robust} show that NMT systems are extremely sensitive to homophone noises and propose joint embedding of textual and phonetic information of a word to improve the robustness to homophone noise.


In this paper, we study the robustness of NMT systems from a different perspective. We replace multiple words in the source sentence while trying to ensure that the predicted translation is unchanged. An NMT system is expected to output a different translation when replacement of multiple words completely changes the semantics of the original source sentence. This is different from prior works where one expects the NMT system to output same (or similar) translation for the noisy source sentence. To replace multiple words, we propose an invariance based adversarial attack. The task of multiple replacements can be broken down into two subtasks: (i) traversing the position indices for replacement (i.e. the order in which words are replaced) and (ii) replacing the word. We propose novel strategies for each of the two subtasks and show that they outperform baseline methods. The experiments are conducted on two language pairs namely English-German (en-de) and English-French (en-fr) and two state-of-the-art NMT systems,  BLSTM-based encoder-decoder with attention~\cite{D15-1166} and Transformer~\cite{transformer}.

\subsection{Contributions of this work}


The main contributions of this paper are summarized below:
\begin{enumerate}
    \item We show that the current state-of-the-art NMT systems are indeed non-robust to multiple word replacements. This shows that the current NMT systems are unable to capture the semantics of the source language.
    \item We propose a novel technique to traverse the position indices for replacement. The proposed technique uses the norm of the gradient of loss with respect to input embeddings. The results show that the proposed technique outperforms random baseline.
    \item Given the traversal of position indices, we propose a soft-attention based technique for choosing a word. The results show that the proposed technique outperforms HotFlip for all the experimental settings by a significant margin.
    \item We propose a BLEU-based metric to evaluate the effectiveness of an invariance-based attack and show the merits of the proposed metric in comparison to success rate. 
\end{enumerate}

\section{Method}

In this section, we describe the proposed method in detail. In Section~\ref{prune}, we outline the vocabulary pruning method which is a pre-processing step of the proposed method. In Section~\ref{sec-Min-Grad}, we describe the proposed technique for position indices traversal. In Section~\ref{sec-Soft-Att}, we describe the proposed technique for word replacement. Finally in Section~\ref{final method}, we describe how the two techniques are used for doing multiple replacements over the source sentence. 

\subsection{Vocabulary Pruning}\label{prune}

The NMT models in the present work use a shared vocabulary for source and target languages. Let $V_{shared}$ denote the shared vocabulary set. We use the training corpus in the source language to find the set of unique words. Let $V_{unique}$ denote this set. We consider the set intersection $V = V_{shared}\cap V_{unique}$. Hence $V$ denotes the set of \emph{proper words} in the source language present in the vocabulary of NMT model. Let $s^{org}=(s_{1}^{org}, s_{2}^{org},..., s_{n}^{org})$ denote  the original sentence in the source language. Given $s^{org}$, we remove the words present in the original sentence from the set $V$, i.e. $V_{prune} = V\setminus s^{org}$. We use $V_{prune}$ to select new words for replacement. 

\subsection{Position Indices Traversal}\label{sec-Min-Grad}

Let $s=(s_{1}, s_{2},..., s_{n})$ denote a sentence in the source language and  $x$ denote the one-hot representation of the sentence $s$ i.e.  $x = ((x_{11},...,x_{1\mid V_{shared}\mid}),...,(x_{n1},...,x_{n\mid V_{shared}\mid}))$ where $x_{ij}$ is $1$ if $j^{th}$ word is present in $i^{th}$ position and $0$ otherwise. Let $e=(e_{1}, e_{2},..., e_{n})$ denote the embedded version of input $x$ where $e_{i}$'s are $d$-dimensional and $t^{org}=(t_{1}^{org}, t_{2}^{org},..., t_{m}^{org})$ denote the predicted translation of the NMT model for the original source sentence $s^{org}$. We consider the standard negative log likelihood loss $L_{nll}$ given by

\begin{equation}
    L_{nll} = - \sum_{i=1}^{m} log(q(t_{i}^{org}|t_{<i}^{org},x))
\end{equation}

where $q(t_{i}^{org}|t_{<i}^{org},x)$ denotes the probability assigned to the word $t_{i}^{org}$ by the NMT model and x is one-hot representation of  the source sentence $s$. Let $ind_{vis} \subseteq \{1,2,...,n\}$ denote the set of position indices which have already been traversed. We choose the position for replacement, $r$, using the following equation

\begin{equation}
    r = \argmin_{i\notin ind_{vis}}\left\|\nabla_{e_{i}} L_{nll}\right\|_{2}
\end{equation}
where $\left\|.\right\|_{2}$ is the $\ell_{2}$-norm, $e_{i}$ is the $i^{th}$ embedding and $\nabla_{e_{i}} L_{nll}$ is the gradient of the loss function with respect to $e_{i}$. The rationale behind choosing the replacement position in this way is that the term $\left\|\nabla_{e_{i}} L_{nll}\right\|_{2}$ tells us about the sensitivity of loss function with respect to the $i^{th}$ embedding $e_{i}$ and hence changing a word at a position which has the minimum $\ell_{2}$-norm should not have a large impact on the predicted translation. We refer to this technique as \textbf{Min-Grad}. We summarize the Min-Grad method in Algorithm~\ref{Min-Grad}.

\begin{algorithm}[htbp]
\caption{Min-Grad}\label{Min-Grad}
\DontPrintSemicolon
  $\textbf{Input: }s,t^{org},ind_{vis}$\;
  $\textbf{Output: }r$\;
  $\text{Get } e, x \text{ from }s$\;
  $\text{Compute loss } L_{nll} \text{ for } (x,t^{org})$\;
  $r = \argmin_{i\notin ind_{vis}}\left\|\nabla_{e_{i}} L_{nll}\right\|_{2}$\;
  \KwRet{$r$}\;
\end{algorithm}

\subsection{Word Replacement}\label{sec-Soft-Att}

Let $r$ denote the position for word replacement. We replace $(x_{r1},x_{r2},...,x_{r\mid V_{shared}\mid})$ with a probability distribution $p$ i.e. $p=(p_{1},p_{2},...,p_{\mid V_{shared}\mid})$ where $p_{i}$ is set to $0$ if the $i^{th}$ word does not belong to $V_{prune}$. We set all the other $p_{i}$'s to be equal initially. Let $x'$ denote the modified input. We modify the non-zero $p_{i}$'s using gradient descent in order to minimize $L_{nll}$. Note that only the non-zero $p_{i}$'s are modified. We modify $p_{i}$'s until either $max_{iter}$ iterations is reached or a particular word is assigned a probability greater than $max_{prob}$ for $n_{iter}$ consecutive iterations. Finally, for the position $r$, we choose the $j^{th}$ word where $j=argmax(p)$. Since this technique picks a word using soft-attention over the vocabulary set $V_{prune}$, we refer to it as \textbf{Soft-Att}. We summarize the Soft-Att method in Algorithm~\ref{Soft-Att}.

\begin{algorithm}[h]
\caption{Soft-Att}\label{Soft-Att}
\DontPrintSemicolon
  $\textbf{Input: }s,t^{org},r$\;
  $\textbf{Output: }ind_{word},loss$\;
  $\text{Initialize } p,x' \text{ using }s,r$\;
  $count = \{\}$\;
  $\text{Initialize } count \text{ to }0 \text{ for all word indices}$\;
  \For{$j\leftarrow 1$ \KwTo $max_{iter}$}
  {
  $loss \gets L_{nll} \text{ for } (x',t^{org})$\;
  $\text{Update }p_{i}\text{'s using gradient descent}$\;
  $\text{Get }x'\text{ from }p$\;
  $p_{max},ind_{word} \gets max(p),argmax(p)$\;
  \For{\text{ind in word indices}}
  {\uIf {$ind \neq ind_{word}$}
  {$count[ind] = 0$\;
  }
  }
  \uIf {$p_{max}>max_{prob}$}
  {$count[ind_{word}] \mathrel{+}= 1$\;
  \uIf {$count[ind_{word}]==n_{iter}$}
  {
  $\textbf{break}$\;
  }
  }
  \uElse{$count[ind_{word}]=0$\;}
  }
  \KwRet{$ind_{word},loss$}\;
\end{algorithm}

\subsection{Proposed method}\label{final method}

In order to make multiple replacements over the original source sentence, $s^{org}$, we use the two methods (Min-Grad and Soft-Att) iteratively. We name the proposed method \textbf{Min-Grad + Soft-Att}.  

\begin{algorithm}[t!]
\caption{Min-Grad + Soft-Att}\label{Min-Grad+Soft-Att}
\DontPrintSemicolon
  $\textbf{Input: }s^{org},t^{org}$\;
  $\textbf{Output: }s^{adv}$\;
  $\text{Get } x \text{ from }s^{org}$\;
  $l_{org} \gets L_{nll} \text{ for } (x,t^{org})$\;
  $n \gets len(s^{org})$\;
  $s \gets s^{org}$\;
  $l_{min} \gets 100$\;
  $ind_{rep} \gets [\text{ }]$\;
  \For{$j\leftarrow 1$ \KwTo $max_{sweep}$}
  {
  $flag \gets \textbf{False}$\;
  $ind_{vis} \gets [\text{ }]$\;
  \While{$len(ind_{vis})\neq n$}
  { 
  $\text{Get } x \text{ from }s$\;
  $l \gets L_{nll} \text{ for } (x,t^{org})$\;
  $r \gets \text{Min-Grad}(s,t^{org},ind_{vis})$\;
  $\textbf{append }r \text{ to } ind_{vis}$\;
  $ind_{word},loss \gets \text{Soft-Att}(s,t^{org},r)$\;
  \uIf {$r \in ind_{rep} \textbf{ and } loss<l$}
  {$l_{min}\gets max(loss,l_{org})$\;
  $s[r]\gets V_{shared}[ind_{word}]$\;
  $flag \gets \textbf{True}$\;}
  \If {$r \notin ind_{rep} \textbf{ and } loss<l_{min}$}
  {$\textbf{append }r \text{ to } ind_{rep}$\;
  $l_{min}\gets max(loss,l_{org})$\;
  $s[r]\gets V_{shared}[ind_{word}]$\;
  $flag \gets \textbf{True}$\;}
  }
  \If{\textbf{not }flag}
  {$\textbf{break}$\;}
  }
  $s^{adv} \gets s$\;
  \KwRet{$s^{adv}$}\;
\end{algorithm}

The proposed method makes at most $max_{sweep}$ sweeps over the source sentence. Within a particular sweep, we choose the position of replacement using Min-Grad method. This is followed by Soft-Att method to identify the new word to replace with, at the particular position. Note that Soft-Att always picks a word from the pruned vocabulary set, $V_{prune}$. Whether the replacement does take place depends on the \emph{min loss criteria}. We initially set the min loss, $l_{min}$, to a very high value (i.e. $100$). This ensures that at least one replacement always takes place. If in a previous sweep, a replacement has taken place at the position identified by the Min-Grad, then we compare the loss obtained from the Soft-Att method with the loss of the current sentence. If the loss obtained from the Soft-Att method is less than the loss of the current sentence, then the replacement is done and $l_{min}$ is updated accordingly. The logic behind this step is to ensure that the new source sentence is better than the old one in terms of $L_{nll}$. Whereas, if no replacement has taken place so far at the position identified by the Min-Grad, then we compare the loss obtained from the Soft-Att method with $l_{min}$. If the loss obtained from the Soft-Att method is less than $l_{min}$, then the replacement is done and $l_{min}$ is updated accordingly. We update 
$l_{min}$ as $l_{min} = max(loss,l_{org})$ where $loss,l_{org}$ are the loss obtained from the Soft-Att method and the original loss respectively. Capping the min loss at original loss allows us to do more replacements while ensuring an optimal solution at the same time. We stop the algorithm if no replacement takes place in a particular sweep. For ease of understanding, we summarize the proposed method in Algorithm~\ref{Min-Grad+Soft-Att}.

Apart from the proposed method, we also study three baseline methods, namely, random + Soft-Att, Min-Grad + HotFlip and random + HotFlip. The \emph{random} baselines refer to the method where traversal of position indices is done randomly instead of via Min-Grad and \emph{HotFlip} baselines refer to the method where word replacement is done via HotFlip instead of Soft-Att. Note that the other methods like ~\cite{cheng-etal-2018-towards, cheng-etal-2019-robust, liu-etal-2019-robust} study robustness of NMT systems in a different framework and hence, these methods are not applicable for comparison with the method presented here. HotFlip being a general method for word/character replacement is relevant to our setting and hence, comparable to the proposed method.  

\section{Implementation Details}


We perform experiments on two language pairs from TED talks dataset~\cite{ted}. The two language pairs are (i) English-German (en-de) and (ii) English-French (en-fr). The dataset statistics for the two language pairs are given in Table~\ref{table:ted-data}. We train BLSTM-based encoder-decoder with attention translation model using OpenNMT-py  for the two language pairs. We use the standard implementation provided in the repository\footnote{\url{https://github.com/OpenNMT/OpenNMT-py}} for training. The model uses attention mechanism proposed by Luong \emph{et al.}\cite{D15-1166}. 

We use the \emph{Transformer base model} configuration~\cite{transformer} for both the language pairs. The model consists of $6$ encoder-decoder layers. We closely follow the implementation provided by Sachan and Neubig ~\cite{devendra2018multilingual} for training the Transformer models. Both the NMT models, BLSTM-based encoder-decoder with attention and Transformer, use byte pair encoding with $32,000$ merge operations~\cite{bpe}. Also, both the NMT models use beam search with beam width of $5$ during prediction. Table~\ref{table:bleu} shows the BLEU score~\cite{bleu} for the trained NMT models on the \emph{test set} of TED dataset. The BLEU scores for Transformer are similar to the results reported by Sachan and Neubig ~\cite{devendra2018multilingual}. As expected, Transformer achieves a higher BLEU score than BLSTM-based encoder-decoder with attention for the two language pairs. 

\begin{table}[t!]
  \begin{center}
    \begin{tabular}{c|l|l|l} 
      \textbf{Language Pair} & \textbf{Training } & \textbf{Dev} &  \textbf{Test}\\ \hline
      en-de & 167,888 & 4,148 & 4,491\\
      en-fr & 192,304 & 4,320 & 4,866\\
    \end{tabular}
  \end{center}
  \caption{Dataset Statistics}
  \label{table:ted-data}
\end{table}

\begin{table}[t!]
  \begin{center}
    \begin{tabular}{c|l|l} 
      \textbf{Model} & \textbf{en-de} & \textbf{en-fr}\\ \hline
      BLSTM & 26.33 & 39.32\\
      Transformer & 29.27 & 43.15\\
    \end{tabular}
  \end{center}
  \caption{BLEU score on the \emph{test set}}
  \label{table:bleu}
\end{table}

\begin{table*}
\begin{center}
\small
\begin{tabular}{c|c|c|c|c|c}
\multirow{2}{*}{\textbf{Model}}      & \multirow{2}{*}{\textbf{Method}}                           & \multicolumn{2}{c|}{\textbf{en-de}}  & \multicolumn{2}{c}{\textbf{en-fr}} \\ \cline{3-6}
                    &                       & \textbf{Success Rate}   & \textbf{NOR}  & \textbf{Success Rate}   & \textbf{NOR}
                   \\ \hline
\multirow{4}{*}{BLSTM}          & random + HotFlip    &   25.4\%      &  0.23, 0.21   &       28.2\%            &   0.21, 0.18                     \\\cline{2-6}
                                & Min-Grad + HotFlip  &   31.8\%      &  0.22, 0.19   &       40.2\%            &   0.19, 0.17
                   \\\cline{2-6}
                                & random + Soft-Att   &   61.2\%      &  0.58, 0.62   &       64.6\%            &   0.62, 0.67
                   \\\cline{2-6}
                                & Min-Grad + Soft-Att &\textbf{67.8\%}&  0.58, 0.61   &    \textbf{70.8\%}      &   0.61, 0.66
                   \\\hline\hline
\multirow{4}{*}{Transformer}    & random + HotFlip    &   35.0\%      &  0.26, 0.24   &       40.6\%            &   0.24, 0.21                     \\\cline{2-6}
                                & Min-Grad + HotFlip  &   45.0\%      &  0.26, 0.24   &       44.0\%            &   0.23, 0.21
                   \\\cline{2-6}
                                & random + Soft-Att   &   50.2\%      &  0.40, 0.39   &       59.0\%            &   0.37, 0.35
                   \\\cline{2-6}
                                & Min-Grad + Soft-Att &\textbf{61.6\%}&  0.41, 0.42   &    \textbf{64.8\%}      &   0.36, 0.34
                   \\\hline
\end{tabular}
\end{center}
\caption{Success rate (in \%) and number of replacements for different methods. \emph{NOR} represents the mean/median of the normalized \textbf{N}umber \textbf{O}f \textbf{R}eplacements across all the sentences. The highest Success rate is marked in bold.}
\label{table:blstm_transformer_success}
\end{table*}

To study the proposed attack, we randomly select $500$ sentences from the \emph{test set} of TED dataset. The values of the different hyperparameters are as follows: $n_{sweep}=5, max_{iter} = 1000, max_{prob} = 0.9$ and $n_{iter} = 10$. The size of the vocabulary set $V$ (i.e. the set of \emph{proper words} in the source language) for English-German and English-French are $9,723$ and $11,699$ respectively. The code for the proposed attack will be made publicly available.

\section{Results}

In this section, we discuss the results of the proposed method in comparison with the baseline methods. In Section~\ref{success-rate}, we look at the success rate and number of replacements of different methods across NMT models. In Section~\ref{bleu-metric}, we evaluate the effectiveness of various method based on a BLEU-based metric. We also argue why the BLEU-based metric is more appropriate than success rate to measure effectiveness of invariance-based attacks. Note that, we use BLSTM as a shorthand for BLSTM-based encoder-decoder with attention in this section. Finally, we try to analyze the nature of replacements in successful adversarial examples and Section~\ref{comment} presents our observations.  

\begin{table*}
\begin{center}
\small
\begin{tabular}{c|c|c|c|c|c|c}
\textbf{Transformer}          & \textbf{Method}      & src   &$l_{1}$&$l_{2}$&$l_{1}^{blstm}$&$l_{2}^{blstm}$ \\ \hline
\multirow{4}{*}{\textbf{en-de}} & random + HotFlip     & 51.04 & 80.49 & 47.53 & 36.42 & 43.66  \\ 
                                & Min-Grad + HotFlip   & 53.23 & 83.13 & 49.15 & 36.51 & 44.76  \\ 
                                & random + Soft-Att    & 32.01 & 84.79 & \textbf{29.72} & \textbf{20.62} & 27.85  \\
                                & Min-Grad + Soft-Att  & \textbf{31.17} & \textbf{88.55} & 31.09 & 20.63 & \textbf{27.43}  \\ \hline
\multirow{4}{*}{\textbf{en-fr}} & random + HotFlip     & 55.51 & 85.18 & 40.35 & 52.00 & 36.18  \\ 
                                & Min-Grad + HotFlip   & 57.92 & 88.40 & 41.98 & 54.39 & 37.68  \\ 
                                & random + Soft-Att    & \textbf{33.61} & 89.77 & \textbf{21.59} & \textbf{32.37} & \textbf{19.09}  \\
                                & Min-Grad + Soft-Att  & 35.40 & \textbf{91.99} & 23.28 & 34.32 & 20.29  \\ \hline
\end{tabular}
\end{center}
\caption{BLEU scores for the original/adversarial sentence (src) and their respective translation by the four NMT models. $l_{1}$ denotes the model under attack, $l_{2}$ denotes the other Transformer model. $l_{1}^{blstm},l_{2}^{blstm}$ are the BLSTM counterparts of $l_{1}$ and $l_{2}$.}
\label{table:transformer_bleu}
\end{table*}

\begin{table*}
\begin{center}
\small
\begin{tabular}{c|c|c|c|c|c|c}
\textbf{BLSTM}          & \textbf{Method}      & src   &$l_{1}$&$l_{2}$&$l_{1}^{trans}$&$l_{2}^{trans}$ \\ \hline
\multirow{4}{*}{\textbf{en-de}} & random + HotFlip     & 57.09 & 71.35 & 48.84 & 43.90 & 49.42  \\ 
                                & Min-Grad + HotFlip   & 59.28 & 75.55 & 50.38 & 45.96 & 52.26  \\ 
                                & random + Soft-Att    & \textbf{13.77} & 87.14 & \textbf{19.20} & \textbf{18.36} & \textbf{21.62}  \\
                                & Min-Grad + Soft-Att  & 14.49 & \textbf{89.86} & 19.74 & 18.51 & 21.98  \\ \hline
\multirow{4}{*}{\textbf{en-fr}} & random + HotFlip     & 60.87 & 79.62 & 39.28 & 58.60 & 41.73  \\ 
                                & Min-Grad + HotFlip   & 63.97 & 84.87 & 41.16 & 61.80 & 44.94  \\ 
                                & random + Soft-Att    & 12.99 & 92.44 & 10.62 & 28.34 & 12.12  \\
                                & Min-Grad + Soft-Att  & \textbf{12.66} & \textbf{93.87} & \textbf{9.95}  & \textbf{27.21} & \textbf{11.92}  \\ \hline
\end{tabular}
\end{center}
\caption{BLEU scores for the original/adversarial sentence (src) and their respective translation by the four NMT models. $l_{1}$ denotes the model under attack, $l_{2}$ denotes the other BLSTM model. $l_{1}^{trans},l_{2}^{trans}$ are the Transformer counterparts of $l_{1}$ and $l_{2}$.}
\label{table:blstm_bleu}
\end{table*}

\subsection{Success rate}\label{success-rate}
Table~\ref{table:blstm_transformer_success} shows the success rate and the mean, median of the number of replacements (normalized by the length of original sentence) for different methods. For a particular NMT model, we define the \emph{success rate} of a method as the percentage of adversarial sentences which were assigned the \emph{same translation} as the original source sentence ($s^{org}$) by the NMT model. We report both the success rate and number of replacements since for two attacks with similar success rate, the one with more number of replacement is better. Furthermore, it is more likely that the meaning of the sentence has changed if the number of replacements are higher. 

1: Comparing Min-Grad and random: As we can see from Table~\ref{table:blstm_transformer_success}, for both Hotflip and Soft-Att, Min-Grad method gives significant improvement in success rate in comparison with random baseline across all the NMT models. The number of replacement for Min-Grad is comparable with random. This shows that the improvement in success rate is significant since otherwise, an attack method can achieve higher success rate by doing less number of replacements. 

2: Comparing Soft-Att and HotFlip: From Table~\ref{table:blstm_transformer_success}, across all the NMT models, we can see that Soft-Att significantly outperforms HotFlip both in terms of success rate and number of replacements. 

3: Comparing BLSTM and Transformer: Table~\ref{table:blstm_transformer_success} shows that Transformer is more robust to our proposed method than BLSTM. This is because our proposed method has less number of replacements and lower success rate in case of Transformer than BLSTM for both the language pairs. Interestingly, HotFlip has higher success rate and similar number of replacement in case for Transformer than BLSTM.

Overall, as is evident from Table~\ref{table:blstm_transformer_success}, our proposed method (Min-Grad + Soft-Att) achieves the highest success rate across the NMT models. 
\begin{table*}[h]
\begin{center}
\small
\begin{tabular}{c|c|c|c}
\multirow{2}{*}{\textbf{Model}}      & \multirow{2}{*}{\textbf{Method(M)}}                           & \multicolumn{2}{c}{\textbf{e(M)}} \\ \cline{3-4}
                    &                       & \textbf{en-de}   & \textbf{en-fr}  
                   \\ \hline
\multirow{4}{*}{BLSTM}          & random + HotFlip    &   45.58      &  44.17                               \\\cline{2-4}
                                & Min-Grad + HotFlip  &   46.46      &  45.40        
                   \\\cline{2-4}
                                & random + Soft-Att   &   17.16      &  14.33   
                   \\\cline{2-4}
                                & Min-Grad + Soft-Att &   \textbf{16.97}      &  \textbf{13.57} 
                   \\\hline\hline
\multirow{4}{*}{Transformer}    & random + HotFlip    &   39.63      &  39.77                             \\\cline{2-4}
                                & Min-Grad + HotFlip  &   40.10      &  40.71   
                   \\\cline{2-4}
                                & random + Soft-Att   &   25.08      &  \textbf{23.38}   
                   \\\cline{2-4}
                                & Min-Grad + Soft-Att &   \textbf{24.35}      &  24.26 
                   \\\hline
\end{tabular}
\end{center}
\caption{$e(M)$ for different methods M (lower values of $e(M)$ imply better attack efficiency).}
\label{table:e(M)}
\end{table*}

\subsection{BLEU-based metric}\label{bleu-metric}

While success rate is the most straightforward metric to measure the efficiency of an invariance based attack on an NMT system, it does have some disadvantages. As mentioned earlier, an attack method can achieve a higher success rate by doing fewer replacements. Hence, comparing both the success rate and number of replacement simultaneously is a better approach. However, there are still few issues that we need to address (a) Although more number of replacements does increase the chances of the meaning of the original sentence being changed, one can think of \emph{pathological examples} where many replacements are made without significant change in the meaning. (b) It is possible that the original/adversarial sentence are assigned the same translation by the NMT model due to the property of the target language rather than a deficiency in the NMT model. As an example, if the target language does not have \emph{gender markers} and \emph{continuous tense}, then the two sentences ``He is playing guitar." and ``She plays guitar." will have the same translation. 


To address these issues, we propose a BLEU-based metric to evaluate efficiency of an invariance based attack. In the present work, we have $4$ NMT models, $2$ for each language pair. Consider the case where Min-Grad+Soft-Att is used to attack \emph{en-de Transformer} resulting in pair of original/adversarial source sentences. To address issue (a), we can translate the original/adversarial source sentences in French using \emph{en-fr Transformer}. If the meaning has not changed significantly, we can expect the BLEU score for the French translations to be high. To address issue (b), we can translate the original/adversarial source sentences in German using \emph{en-de BLSTM} (since the target language is German). If the translations of Transformer were similar due to the property of target language, we can expect the BLEU score for the German translations by the BLSTM to be high as well.

To summarize, an \emph{effective} invariance based attack is expected to give pair of original/adversarial source sentences whose corresponding translations by the model under attack have \emph{high} BLEU scores and whose corresponding translations by the other NMT models have \emph{low} BLEU scores.

Table~\ref{table:transformer_bleu} shows the BLEU scores for the original/adversarial sentence (src) and their respective translation by the four NMT models. In Table~\ref{table:transformer_bleu}, $l_{1}$ denotes the \emph{Transformer} model under attack (e.g.\ en-de), $l_{2}$ denotes the other Transformer model (e.g.\ en-fr), and $l_{1}^{blstm},l_{2}^{blstm}$ are the BLSTM counterparts of $l_{1}$ and $l_{2}$. Similarly, Table~\ref{table:blstm_bleu} shows the BLEU scores for the original/adversarial sentence (src) and their respective translation by the four NMT models. In Table~\ref{table:blstm_bleu}, $l_{1}$ denotes the \emph{BLSTM} model under attack, $l_{2}$ denotes the other BLSTM model, and $l_{1}^{trans},l_{2}^{trans}$ are the Transformer counterparts of $l_{1}$ and $l_{2}$. For an attack to be effective,  BLEU score for $l_{1}$ should be high and the other four BLEU scores should be low. Note that the BLEU score for $src$ is related with the number of replacements reported in Table~\ref{table:blstm_transformer_success}. The two metrics are inversely related; more number of replacement implies lower BLEU score for $src$. 

From Tables~\ref{table:transformer_bleu} and~\ref{table:blstm_bleu}, we can see that Soft-Att achieves a higher BLEU score for $l_{1}$ in comparison with HotFlip for all the experimental settings. Moreover, the other four BLEU score are lower for Soft-Att than HotFlip. This result showcases the efficiency of the proposed method since it outperforms HotFlip in terms of success rate, number of replacements and BLEU scores. The fact that, for the proposed method, BLEU scores is low for other NMT models also shows that the adversarial sentences are not transferable in nature.  In other words, the pair of original/adversarial sentences are specific to the NMT model. 

In a general setting, let there be $n$ NMT models denoted by $l_{1}$,$l_{2}$,..,$l_{n}$ where $l_{1}$ is the NMT model under attack. Using the BLEU-based metric, we propose a composite score, $e(M)$ to evaluate the efficiency of an attack method M as follows.

\begin{equation}
    e(M) = \frac{b_{src}+(100-b_{l_{1}})+...+b_{l_{n}}}{n+1}
\end{equation}

where $b_{src},b_{l_{i}}$ denote the BLEU score for $src$ and NMT model $l_{i}$ respectively. For an attack method M to be more effective, $e(M)$ should be lower. Table~\ref{table:e(M)} shows $e(M)$ values for different methods and across NMT models. This table nicely summarizes the results presented in Tables~\ref{table:transformer_bleu} and~\ref{table:blstm_bleu}. The $e(M)$ values demonstrate that the state-of-the-art NMT systems are unable to capture the semantics of the adversarial examples generated by the the proposed method, Min-Grad+Soft-Att.   

\begin{table*}[h]
\begin{center}
\small
\begin{tabular}{r|l|l}
\hline
\multirow{6}{*}{\textbf{en-de}} 
& \textbf{src}    & And because God loves her , I did get married . \\
& \textbf{adv-src}  & plus because God loves them kilograms me been abused married . \\
& \textbf{pred}     & Und weil Gott sie sie liebt , wurde ich verheiratet . \\
\cline{2-3}
& \textbf{src}    & I want to know the people behind my dinner choices . \\
& \textbf{adv-src}  & I want ordinarily know the humans behind my dinner flog arguments\\
& \textbf{pred}     & Ich m\"ochte die Menschen hinter meinen Abendessen kennen . \\
\hline
\multirow{6}{*}{\textbf{en-fr}}
& \textbf{src}    & I was clearly more nervous than he was . \\
& \textbf{adv-src} & adaptations was clearly more nervous label he was . \\
& \textbf{pred}     & J'\'etais clairement plus nerveux qu'il \'etait . \\
\cline{2-3}
& \textbf{src}    & A dome , one of these ten-foot domes . \\
& \textbf{adv-src} & An dome pale an of Those exes 3 foot domEvelyn tat \\
& \textbf{pred}     & Un d\^ome , un de ces d\^omes de 3 m\`etres . \\
\hline
\end{tabular}
\end{center}
\caption{Examples of \emph{Min-Grad + Soft-Att} for BLSTM-based encoder-decoder with attention. The NMT model predicts the same translation for src and adv-src.} 
\label{table:ex-blstm}
\end{table*}


\begin{table*}
\begin{center}
\small
\begin{tabular}{r|l|l}
\hline
\multirow{6}{*}{\textbf{en-de}} 
& \textbf{src}    & Is it something about the light ? \\
& \textbf{adv-src}  & Is Bald passage about the light ? \\
& \textbf{pred}     & Geht es um das Licht ? \\
\cline{2-3}
& \textbf{src}    & So the whole is literally more than the sum of its parts . \\
& \textbf{adv-src}  & Small the whole is bucks more than number sum Von His parts rank \\
& \textbf{pred}     & Das Ganze ist mehr als die Summe seiner Teile . \\
\hline
\multirow{6}{*}{\textbf{en-fr}}
& \textbf{src}    & They look like the stuff we walk around with . \\
& \textbf{adv-src} & Hudson look like the ping we walk fishes with . \\
& \textbf{pred}     & Ils ressemblent \`a ce que nous marchons avec . \\
\cline{2-3}
& \textbf{src}    & There are many , many problems out there . \\
& \textbf{adv-src} & look numerous supported stays behold problems hundred there . \\
& \textbf{pred}     & Il y a de nombreux probl\`emes . \\
\hline
\end{tabular}
\end{center}
\caption{Examples of \emph{Min-Grad + Soft-Att} for Transformer. The NMT model predicts the same translation for src and adv-src.} 
\label{table:ex-transformer}
\end{table*}

\subsection{A Comment on Types of Words Replaced}\label{comment} 
In order to understand what types of words are replaced to generate successful adversarial examples, we observe that there is no clear trend about the types of words replaced. Both highly frequent (stop words) and thematic words are getting replaced. The model under attack remains invariant to replacement of highly thematic words as well as frequent words by semantically very different words. Invariance is observed even in case of introduction of named-entities (NEs). While trying to understand if specific parts-of-speech (POS) are vulnerable, no clear tendency is noted. These observations are highlighted through the examples given in Tables~\ref{table:ex-blstm} (for BLSTM-based encoder-decoder with attention model) and~\ref{table:ex-transformer} (for Transformer based translation model) which are generated by \emph{Min-Grad+Soft-Att} method.

\section{Conclusion and Future Work}
The proposed study shows word-level \emph{undesirable invariances} captured by an NMT system. We define \emph{undesirable invariances} as the scenario in which the predicted translation remains unchanged when multiple words in the source sentence are replaced changing the semantic of the input sentence. Two language pairs, namely, English-German (en-de) and English-French (en-fr) are considered to investigate the behaviour of two state-of-the-art NMT systems: BLSTM-based encoder-decoder with attention and Transformer. We break down the problem of replacing a word into two sub-problems: traversing position indices and replacing a word given a position. Two techniques, \emph{Min-Grad} and \emph{Soft-Att} are proposed for the two sub-problems. The results show that the proposed techniques significantly outperform HotFlip and random related baselines. We also propose an alternate BLEU-based metric to evaluate the effectiveness of an invariance based attack and argue its effectiveness in comparison to success rate. 

This study was motivated to explore the robustness of NMT systems to nonsensical inputs. Our results demonstrate that although the state-of-the-art NMT systems achieve high BLEU score, these systems are not efficient enough to capture the semantics of the source sentence. This shows the need to enhance robustness of NMT systems to nonsensical inputs. In the works of \cite{cheng-etal-2018-towards,cheng-etal-2019-robust,liu-etal-2019-robust}, the authors modify the training algorithms to improve robustness of NMT systems to the particular type of noise in consideration. Similar approach has been followed to develop robust image classifiers \cite{Madry}. However, for the proposed attack, developing a noise-aware training algorithm is a challenging task. This is due to the lack of \emph{gold translations} for the adversarial examples obtained via the proposed attack which was not the case for the previous studies. Hence, developing robust training strategies to counter invariance-based attacks is a possible area of future research.



\bibliographystyle{IEEEtran}








\end{document}